\documentclass[runningheads]{llncs}
\usepackage[dvipsnames]{xcolor}
\usepackage{graphicx}
\usepackage{comment}
\usepackage{amsmath,amssymb} % define this before the line numbering.
\usepackage{color}

\usepackage{caption}
\usepackage{makecell}
\usepackage{subcaption}
\usepackage{enumitem}
\usepackage{array}
\usepackage{tikz}
\usepackage{pgfplots}
\usepackage{sansmath}
\usepackage{multirow}
\usepackage{booktabs}
\usepackage{cite}
\usepackage{pifont}
\usepackage[dvipsnames]{xcolor}
\usepackage{bm}
\usepackage{mathrsfs}

\newcolumntype{x}[1]{>{\centering\arraybackslash\hspace{0pt}}p{#1}}

\usepackage[accsupp]{axessibility}  % Improves PDF readability for those with disabilities.

\begin{document}
% \linenumbers
\pagestyle{headings}
\mainmatter
\def\ECCVSubNumber{4136}  % Insert your submission number here

%\title{Understanding long movie videos with multi-scale spacetime state-space model} % Replace with your title
\title{Long Movie Clip Classification with \\  State-Space Video Models}

%\begin{comment}
% \titlerunning{ECCV-22 submission ID \ECCVSubNumber} 
% \authorrunning{ECCV-22 submission ID \ECCVSubNumber} 
% \author{Anonymous ECCV submission}
% \institute{Paper ID \ECCVSubNumber}

\titlerunning{ViS4mer} 
\authorrunning{} 
\author{Md Mohaiminul Islam, Gedas Bertasius}
\institute{UNC Chapel Hill}

%\end{comment}
%******************

\begin{comment}
\titlerunning{Abbreviated paper title}
\author{First Author\inst{1}\orcidID{0000-1111-2222-3333} \and
Second Author\inst{2,3}\orcidID{1111-2222-3333-4444} \and
Third Author\inst{3}\orcidID{2222--3333-4444-5555}}
%
\authorrunning{F. Author et al.}
\institute{Princeton University, Princeton NJ 08544, USA \and
Springer Heidelberg, Tiergartenstr. 17, 69121 Heidelberg, Germany
\email{lncs@springer.com}\\
\url{http://www.springer.com/gp/computer-science/lncs} \and
ABC Institute, Rupert-Karls-University Heidelberg, Heidelberg, Germany\\
\email{\{abc,lncs\}@uni-heidelberg.de}}
\end{comment}
%******************
\maketitle

\begin{abstract}

Most modern video recognition models are designed to operate on short video clips (e.g., 5-10s in length). Thus, it is challenging to apply such models to long movie understanding tasks, which typically require sophisticated long-range temporal reasoning. The recently introduced video transformers partially address this issue by using long-range temporal self-attention. However, due to the quadratic cost of self-attention, such models are often costly and impractical to use. Instead, we propose ViS4mer, an efficient long-range video model that combines the strengths of self-attention and the recently introduced structured state-space sequence (S4) layer. Our model uses a standard Transformer encoder for short-range spatiotemporal feature extraction, and a multi-scale temporal S4 decoder for subsequent long-range temporal reasoning. By progressively reducing the spatiotemporal feature resolution and channel dimension at each decoder layer, ViS4mer learns complex long-range spatiotemporal dependencies in a video. Furthermore, ViS4mer is $2.63\times$ faster and requires $8\times$ less GPU memory than the corresponding pure self-attention-based model. Additionally, ViS4mer achieves state-of-the-art results in  $6$ out of $9$ long-form movie video classification tasks on the Long Video Understanding (LVU) benchmark. Furthermore, we show that our approach successfully generalizes to other domains, achieving competitive results on the Breakfast and the COIN procedural activity datasets. The code is publicly available. \footnote{\url{https://github.com/md-mohaiminul/ViS4mer}}

\end{abstract}

\section{Introduction}\label{sec:introduction}

\begin{figure}
    \centering
    \includegraphics[width=\textwidth]{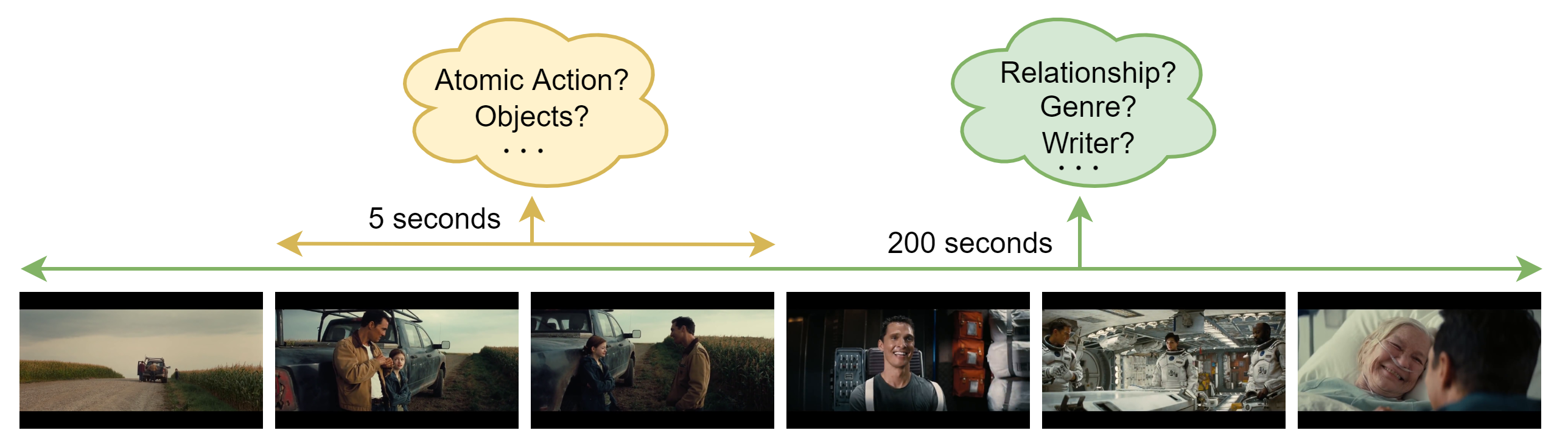}
    \caption{\small Most traditional video models are designed for local prediction tasks (e.g., atomic action recognition, object detection, etc.) in short video clips (e.g., 5 seconds in length). In contrast, we aim to understand complex video understanding tasks in long movie videos (e.g., 200 seconds long), such as classifying the relationships among movie characters, predicting the writer of the story, categorizing the genre of the movie, etc.}
\label{fig:intro}
\end{figure}

Suppose we ask someone to describe the relationship between the characters from the `Interstellar' movie illustrated in Figure~\ref{fig:intro}. It might be difficult for them to do so just by looking at a short video clip of several seconds. However, this is a much easier task if a person watches the whole movie. Thus, in this work, we pose the question of whether we can develop a computer vision model that can leverage long-range temporal cues to answer complex questions such as  `What is the genre of the movie?', `What is the relationship among the characters?', `Who is the director of the movie?', etc. 

The majority of modern video recognition models~\cite{carreira2017quo, feichtenhofer2020x3d, simonyan2014two, feichtenhofer2019slowfast, tran2019video, karpathy2014large, gberta_2021_ICML, arnab2021vivit, liu2021video, fan2021multiscale, patrick2021keeping} are unfortunately not equipped to solve these tasks as they are designed for short-range videos (e.g., 5-10 seconds in duration). Furthermore, extending these models to the long-range video setting by stacking more input video frames is impractical due to excessive computational cost and GPU memory consumption.

Recently, several Transformer models~\cite{gberta_2021_ICML,  wu2021towards} have been shown to perform well on long-range video understanding tasks. However, due to the quadratic cost of standard self-attention, these models are either very computationally costly~\cite{gberta_2021_ICML}, or they have to operate on pre-extracted CNN features~\cite{wu2021towards}, which discard fine-grained spatiotemporal information, thus, limiting the expressivity of the final video model. The latter characteristic is particularly important for long-range movie clip analysis, since fine-grained spatiotemporal cues may be indicative of the relationships between different movie characters, the genre of a movie, etc. 

To address the efficiency-related issues of standard self-attention, recent work in Natural Language Processing (NLP) has proposed a structured state-space sequence model (S4)~\cite{gu2021efficiently} for long-range sequence analysis. Unlike self-attention, the S4 layer has linear memory and computation cost with respect to the input length. As a result, it can handle much longer input sequences. %than the self-attention operator does. 

Combining the strengths of the self-attention and the S4 layer, we propose ViS4mer, a long-range \textbf{Vi}deo classification model composed of a standard transformer encoder and a multi-scale temporal \textbf{S4} decoder. The transformer encoder is used for spatial short-range video feature extraction whereas the S4 decoder performs long-range temporal reasoning, which is necessary for complex movie clip classification tasks. We build our temporal S4 decoder using a multi-scale architecture progressively reducing the number of tokens and the channel dimension with every layer. This allows our model to learn hierarchical spatiotemporal video representation while also reducing the computational cost associated with operating on a large number of video tokens. 

We validate ViS4mer on the recently introduced Long Video Understanding (LVU) benchmark~\cite{wu2021towards}, which consists of nine diverse movie understanding tasks. We show that ViS4mer outperforms previous approaches in $6$ out of $9$ long-range video classification tasks. Moreover, compared to its self-attention counterpart, ViS4mer is $2.63\times$ faster and requires $8\times$ less GPU memory. Lastly, ViS4mer generalizes to other domains, achieving competitive results on the Breakfast~\cite{kuehne2014language} and COIN~\cite{tang2019coin} long-range procedural activity datasets.

\section{Related Work}

\hspace{\parindent}\textbf{Modeling Long Sequences.} Long sequence modeling is a fundamental task in Natural Language Processing (NLP). Previously, Bahdanau \textit{et al.}~\cite{bahdanau2014neural} proposed a recurrent attention scheme for machine translation. Improving upon this work, Vaswani \textit{at el.}~\cite{vaswani2017attention} introduced a self-attention operation for the same machine translation task. Subsequently, a plethora of transformer-based architectures has been proposed for various NLP tasks~\cite{devlin2018bert, liu2019roberta, yang2019xlnet, dai2019transformer, brown2020language, raffel2019exploring}. However, one major drawback of the transformer architecture is the quadratic complexity of standard self-attention. Various efficient self-attention schemes have been proposed for reducing the computation cost when modeling long sequences~\cite{kitaev2020reformer, zaheer2020big, katharopoulos2020transformers, choromanski2020rethinking}. Most relevant to our work, is the method of Gu \textit{at el.}~\cite{gu2021combining, gu2021efficiently, goel2022s} that proposes an efficient structured state-space sequence (S4) layer for long sequence modeling. Inspired by this work, in this paper, we design a video architecture that incorporates the ideas from~\cite{gu2021combining, gu2021efficiently} for long-range movie understanding tasks.

\textbf{Video Recognition.} Most existing video recognition methods are built using 2D and 3D Convolutional Neural Networks~\cite{carreira2017quo, feichtenhofer2020x3d, simonyan2014two, feichtenhofer2019slowfast, tran2019video, karpathy2014large, wu2020multigrid, zhou2018temporal, zolfaghari2018eco, li2020smallbignet, peng2018two}. Due to the local nature of 2D and 3D convolutions, most of these models typically operate on short video clips of a few seconds. Inspired by the success of Transformer models in natural language processing (NLP), recently the transformer-based models have been successfully used for video recognition tasks~\cite{gberta_2021_ICML, arnab2021vivit, liu2021video, fan2021multiscale, patrick2021keeping}. However, due to the quadratic cost of the self-attention operation, these models are very computationally costly and, thus, only applied to short-range video segments. Our work aims to address this issue by proposing a novel efficient ViS4mer model for long movie clip understanding tasks.

\textbf{Understanding Long-form Movie Videos.} Movie understanding is a popular area of video understanding with many prior methods designed for movie-based tasks. Tapaswi \textit{et al.}~\cite{tapaswi2016movieqa} introduce a movie question answering dataset. Bain \textit{et al.}~\cite{bain2020condensed} and Zellers \textit{at al.}~\cite{zellers2019recognition} propose text-to-video retrieval and question answering benchmarks. However, these multi-modal benchmarks are often biased towards the language domain and are not ideal for evaluating video-only models. Several prior works introduced movie understanding datasets ~\cite{xiong2019graph, vicol2018moviegraphs, huang2020movienet}, which are not publicly accessible for copyright issues. Recently, Wu \textit{et al.}~\cite{wu2021towards} introduced a Long-form Video Understanding (LVU) benchmark that uses publicly available MovieClips~\cite{movieclips}. The proposed LVU benchmark contains nine diverse tasks covering a wide range of aspects of long-form video understanding, which makes it suitable for evaluating our work as well. The current state-of-the-art Object Transformer  method~\cite{wu2021towards} applied on this benchmark, uses a Transformer architecture, and a variety of external modules (e.g., short-term video feature extractor~\cite{wang2018non, feichtenhofer2019slowfast}, object detection, and tracking modules~\cite{ren2015faster, lin2017feature, he2017mask}, and self-supervised pretraining) Instead, in this work, we propose ViS4mer, a simple and efficient long-range video recognition model. %that achieves state-of-the-art performance in most LVU benchmark tasks.

\begin{table}[t]
    \centering
    \caption{\small Theoretical runtime and memory requirement of state-space and self-attention operations w.r.t sequence length (L), batch size (B), and hidden dimension (H). Tildes denote log factors~\cite{gu2021efficiently}. The runtime and memory cost of the state-space layer is linear w.r.t the input sequence length as opposed to the quadratic cost of self-attention.}
    \label{tab:theoritical}
    \begin{tabular}{@{\extracolsep{2pt}} lll}
         & Self-attention & State-space \\
         \hline
         Parameters & $\mathrm{H^2}$ & $\mathrm{H^2}$ \\
         Memory & $\mathrm{B(L^2+HL)}$ &  $ \mathrm{BLH}$ \\
         Training & $\mathrm{B(L^2H+LH^2)}$ & $\mathrm{BH(\tilde{H}+\tilde{L})+B\tilde{L}H}$ \\
         Inference & $\mathrm{L^2H+LH^2}$ & $\mathrm{H^2}$\\
         \hline
    \end{tabular}
\end{table}

\section{Background: Structured State-Space Sequence Model}\label{sec:s4_layer}

Before describing our method, we first review some background information on Structured State-Spaces Sequence layers~\cite{gu2021efficiently}, which is one of the key components of our ViS4mer architecture. We start from the fundamental State-Space Model (SSM), which is defined by the simple equation~{(\ref{eq:ssm})}. It maps a 1-dimensional input signal $u(t)$ to an N-dimensional latent space $x(t)$, then projects the hidden state $x(t)$ to a 1-dimensional output signal $y(t)$.
 
\begin{equation}\label{eq:ssm}
\begin{aligned}
  & x'(t) = Ax(t) + Bu(t) \\
  & y(t) = Cx(t) + Du(t)
\end{aligned}
\end{equation}

Here,  A, B, C, and D are parameters learned using gradient descent. Unfortunately, the standard implementation of SSM can be very costly because computing the hidden state requires $L$ successive multiplications with the matrix $A$. This results in $O(N^2L)$ operations and $O(NL)$ space for state dimension N and sequence
length L. Moreover, this operation suffers
from the vanishing/exploding gradients
problem. To address this issue, the recent work~\cite{gu2021efficiently} leverages HiPPO theory~\cite{gu2020hippo}, which requires the $A$ matrix to be defined as: 

\begin{equation}\label{eq:hippo}
A_{nk} = 
\begin{cases}
(2n+1)^{1/2}(2k+1){1/2} & \text{if } n > k\\
n+1 & \text{if } n = k\\
0 & \text{if } n < k
\end{cases}
\end{equation}

This provides theoretical guarantees allowing SSMs to capture long-range dependences in the sequential data. Building on this work, the method in~\cite{gu2021efficiently} develops a structured state-space sequence (S4) layer, which significantly reduces the computation cost of a basic SSM.

In Table~\ref{tab:theoritical}, we compare the theoretical time and space complexity of the self-attention and structured state-space sequence layers. We observe that self-attention has quadratic complexity w.r.t input sequence length $L$ for training time, inference time, and memory requirement. In contrast, the state-space operation has linear time and space dependency w.r.t the input sequence length $L$. We refer the reader to the original paper~\cite{gu2021efficiently} for further details.

\begin{figure}[t]
    \centering
    \includegraphics[width=1\textwidth]{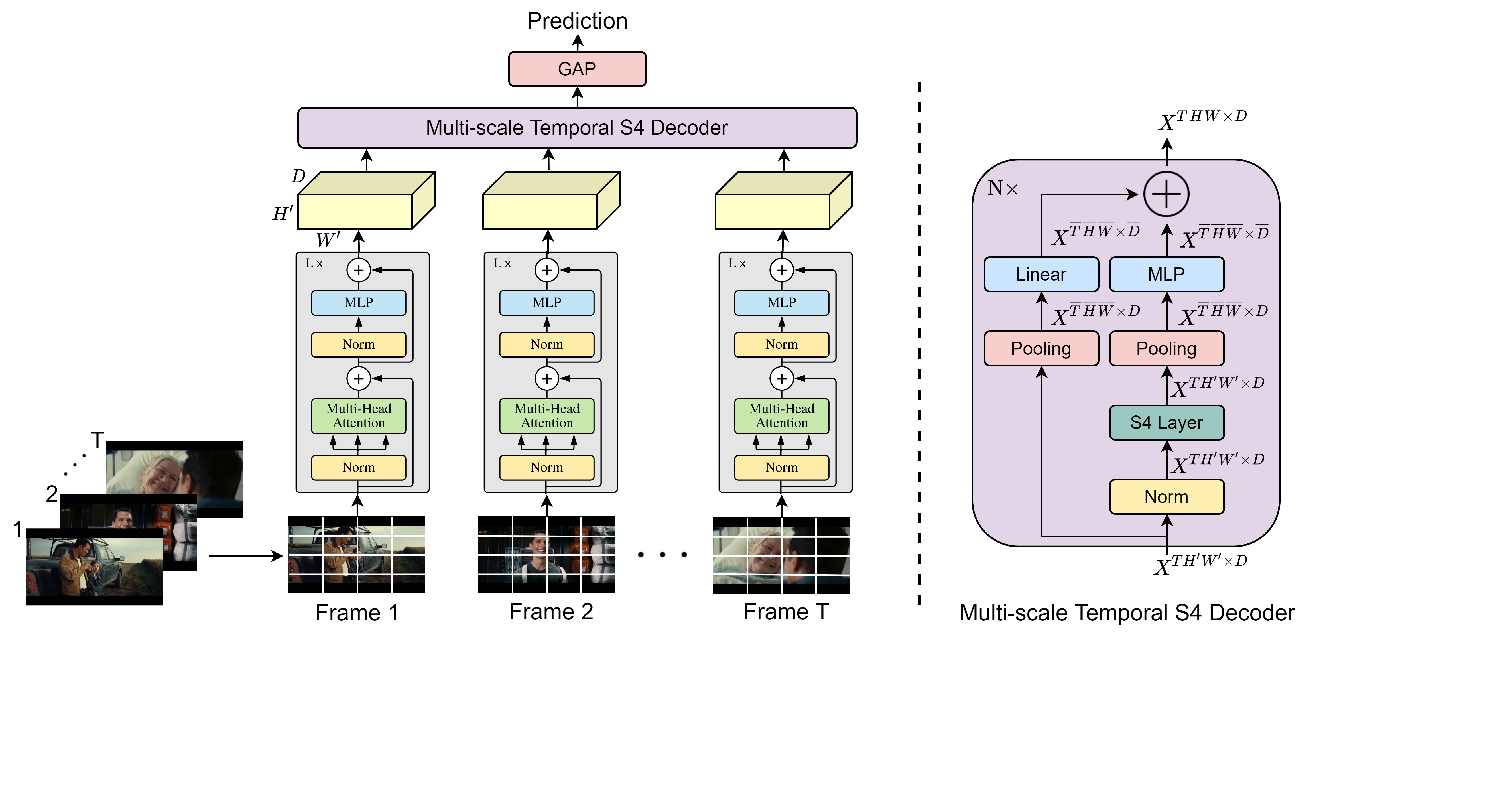}
    \caption{\small Overview of the ViS4mer model. First, we split the video frames into fixed-size patches and use a short-range transformer encoder to extract contextual patch-level features for each video frame. Next, we feed the resulting spatiotemporal patch tokens from the whole video to a novel multi-scale temporal S4 decoder for modeling long-range temporal interactions in movie clips. Each S4 decoder block reduces the spatiotemporal resolution and the channel dimension using a Pooling and an MLP layer for learning multi-scale features. Afterward, the outputs from the S4 decoder are aggregated using global average pooling (GAP), and fed into the classification layer for the final downstream task prediction.}
    \label{fig:ViS4mer}
\end{figure}

\section{The ViS4mer Model}\label{sec:ViS4mer}

Our goal is to design a model for long-range movie clip analysis. To this end, we propose ViS4mer, a long-range video recognition model comprised of a transformer encoder and a multi-scale temporal S4 decoder. Following Vision Transformer~\cite{dosovitskiy2020image}, we first divide input video frames into smaller patches. We then apply a standard transformer encoder to extract fine-grained patch-level features from each video frame. Afterward, we use our proposed multi-scale temporal S4 decoder for long-range temporal reasoning over the patch-level features. Since the decoder has access to fine-grained spatiotemporal video patch information, it can effectively recognize the complex global properties of a long-range video. In Figure~\ref{fig:ViS4mer}, we illustrate the overall architecture of our proposed ViS4mer model. Below, we also discuss each of these components in more detail.

\subsection{Transformer Encoder}\label{transformer}

Let us assume, we have a video $V\in\mathbb{R}^{T\times H\times W\times 3}$ of $T$ frames denoted by $(f_1, ...,f_i, ..., f_T)$. Each frame has a spatial resolution of $H\times W\times 3$, where $H$ is the height, and $W$ is the width of the frame. The transformer encoder $\mathscr{E}$ is then applied to each frame independently.

Following ViT~\cite{dosovitskiy2020image}, we divide each frame into $N$ non-overlapping patches of size $P\times P$, where $N=HW/P$. Then a linear layer is applied to project each patch to a latent dimension of $D$, and a positional embedding $E\in\mathbb{R}^{N\times D}$ is added to each patch embedding. We can think of these embeddings as a sequence of $N$ tokens $(z_1, ..., z_i, ..., z_N)$, where $z_i\in\mathbb{R}^D$.

The resulting sequence is passed through the transformer encoder $\mathscr{E}$ which is a stack of $L$ transformer blocks. Each transformer block contains a multi-headed attention (MHA) and a multi-layer perceptron (MLP) block. Layer normalization (LN) is applied before each block, and a skip connection layer is added after each block. These operations can be expressed as:

\begin{equation}
\begin{aligned}
    &\boldsymbol{z}^\prime = \text{MHA(LN(}\boldsymbol{z}_{in}\text{))} + \boldsymbol{z}_{in}\\
    &\boldsymbol{z}_{out} = \text{MLP(LN(}\boldsymbol{z}^\prime\text{))} + \boldsymbol{z}^\prime\\
\end{aligned}
\end{equation}

The spatiotemporal token outputs of the transformer encoder can then be denoted as $(h_1, ..., h_i, ..., h_T$), where $h_i=\mathscr{E}(f_i)\in\mathbb{R}^{H^\prime \times W^\prime \times D}$, $H^\prime = H/P$, $W^\prime = W/P$. All of these outputs are stacked into a single matrix $X\in\mathbb{R}^{T\times H^\prime \times W^\prime \times D}$.

\subsection{Multi-scale Temporal S4 Decoder}

Next, we describe our framework for modeling complex long-range temporal dependencies among the spatiotemporal tokens produced by the transformer encoder. One obvious choice is to use another transformer model to process such a sequence of tokens. However, this can be challenging due to (i) a large number of spatiotemporal tokens and (ii) the quadratic complexity of the self-attention operation. To illustrate this point, we note that processing a video of $60$ frames of $224\times 224$ spatial resolution using a patch size of $16\times 16$ yields a total of $14\times14\times60=11,760$ output tokens. Using a standard self-attention operator on such a large number of tokens requires $\sim$138 million pairwise comparisons which is extremely costly. Another solution would be to only consider CLS token outputs for each frame. However, doing so removes fine-grained spatiotemporal information, thus, degrading the performance of long-range movie understanding tasks (as shown in our experimental evaluation). 

To overcome this challenge we design a temporal multi-scale S4 decoder architecture for complex long-range reasoning. Instead of using self-attention, our decoder model utilizes the recently introduced S4 layer (described in Section~\ref{sec:s4_layer}). Since the S4 layer has a linear computation and memory dependency with respect to the sequence length, this significantly reduces the computational cost of processing such long sequences.

To effectively adapt the S4 layer to the visual domain, we design a multi-scale S4 decoder architecture. Our temporal multi-scale S4 decoder consists of multiple blocks where each block operates on different spatial resolution and channel dimensions. Starting from a high spatial resolution and high channel dimension, the proposed model gradually decreases spatiotemporal resolution and channel dimension at each block. Our multi-scale architecture is inspired by several successful multi-scale models in the visual domain such as Feature Pyramid Networks~\cite{lin2017feature}, and Swin transformer~\cite{liu2021swin}. Because of this multi-scale strategy, different blocks can effectively learn features at different scales, which helps the model to learn complex spatiotemporal dependencies over long videos. Furthermore, operating on shorter input sequences of smaller channel dimensions in the deeper blocks helps us to reduce overfitting on a relatively small LVU benchmark. Overall, as will be shown in our experimental section, in addition to producing better performance, our multi-scale strategy further reduces the computational cost and GPU memory requirements.

Figure~\ref{fig:ViS4mer} (right) shows the architecture of the multi-scale temporal S4 decoder. The decoder network $\mathscr{D}$ consists of $N$ blocks, which are defined below:
%Let us assume a decoder block receives a spatiotemporal feature $X\in\mathbb{R}^{T\times H' \times W' \times D}$. The operations of a decoder block can then be

\begin{equation}
\begin{aligned}
    &\boldsymbol{x}_{s4} = \text{S4(LN(}\boldsymbol{x}_{in}\text{))}\\
    &\boldsymbol{x}_{mlp} = \text{MLP(Pooling(}\boldsymbol{x}_{s4}\text{))}\\
    &\boldsymbol{x}_{skip} = \text{Linear(Pooling(}\boldsymbol{x}_{in}\text{))}\\
    &\boldsymbol{x}_{out} = \boldsymbol{x}_{mlp} + \boldsymbol{x}_{skip}
\end{aligned}
\end{equation}

%where each of these operations is described in more detail below.

%We describe each of these operations below.

\textbf{S4 Layer.} We flatten the input tensor $X$ to a sequence of $L$ vectors $x_{in} = (x_1, ..., x_i, ..., x_L)$, where $L=T\times H'\times W'$, and $x_i\in\mathbb{R}^{D}$. We then pass this sequence to the S4 layer, which outputs the feature tensor $x_{s4} \in \mathbb{R}^{L \times D}$. %of the same dimensionality. 

\textbf{Spatiotemporal Resolution Reduction.} We reduce the space-time resolution of our input tensor by a factor of $s_T\times s_H\times s_W$ using a max-pooling layer where $s_T\times s_H\times s_W$ is the stride along each axis of the input tensor. The resulting tensor has dimensionality of $\overline{T}\times\overline{H}\times\overline{W}$. This allows our model to learn multi-scale spatiotemporal representations while also reducing the computational cost of operating on long sequences.

\textbf{Channel Reduction.} After the pooling layer, we apply an MLP, which reduces the channel dimension of the input tensor. In addition to decreasing the computational cost, this also reduces overfitting on the LVU benchmark.

\textbf{Skip Connections.} We use skip connections from the input tensor to the final output of a decoder block. Due to the mismatch of feature dimensionalities, we apply an additional pooling layer to the input tensor. Furthermore, to handle the channel dimension mismatch, we use an additional linear layer. %to project the channel dimension of the input tensor to the same dimension as for the final output. 

\subsection{Loss Functions}

 Following~\cite{wu2021towards}, we use a cross-entropy loss for the classification tasks, and the mean squared error (MSE) for the regressions tasks.

\begin{equation}
    L_{ce}(\mathcal{F_C}(\theta)) = -\frac{1}{B}\sum_{i=1}^{B}\sum_{j=1}^K
    y_j^i\log(\mathcal{F_C}(\theta ;x^i)_j)
\end{equation}

\begin{equation}
    L_{mse}(\mathcal{F_R}(\theta)) = -\frac{1}{B}\sum_{i=1}^{B}(y^i-\mathcal{F_R}(\theta ;x^i))^2
\end{equation}

Here, $\mathcal{F_C}(\theta)$ and $\mathcal{F_R}(\theta)$ are our classification and regression models respectively, $B$ is the batch size, $K$ is the number of classes (for the classification task), $y$ is the label, $x$ is the input, and $\theta$ are the learnable model parameters.

\subsection{Implementation Details}\label{sec:implementation}

We resize each video frame to the spatial resolution of $224\times 224$ and use a patch size of $16\times 16$. For the transformer encoder, we use a 24-block transformer with hidden dimension $1024$ pretrained on ImageNet~\cite{deng2009imagenet}. For our multi-scale temporal S4 decoder, we use a $3$-block architecture. Each block has a pooling layer with a kernel of $1 \times 2\times 2$, stride of $1 \times 2\times 2$, and padding of $1 \times 1\times 1$. As discussed above, each block also has an MLP layer, which reduces the feature dimension by a factor of $2\times$. For all of our experiments, we use Adam optimizer~\cite{kingma2014adam} with a learning rate of $10^{-3}$, and with a weight decay of $0.01$. We train our models using NVIDIA RTX A6000 GPU with a batch size of $16$. %We will release our code and pretrained models upon the publication of the paper.

\section{Experiments}

We evaluate ViS4mer on the recently proposed Long-form Video Understanding (LVU) benchmark~\cite{wu2021towards} which contains nine diverse tasks related to long-form movie understanding. Moreover, we also perform thorough ablation studies (i) comparing our S4-based model to an equivalent self-attention baseline, (ii) studying the efficiency of ViS4mer, (iii) comparing our method with other efficient attention schemes, (iv) analyzing the design choices of our model, (v) validating the robustness to different short-range encoders, and (vi) lastly investigating our model's long-range modeling capability. Additionally, to demonstrate the generalization of our approach, we also validate ViS4mer on two long-range procedural activity datasets, COIN~\cite{tang2019coin, tang2020comprehensive} and  Breakfast~\cite{kuehne2014language}. %and show that it performs well in those settings too.

\subsection{Main Results on the LVU benchmark}\label{sec:main result}

The long-form video understanding benchmark (LVU) \cite{wu2021towards} is constructed using the publicly available MovieClip dataset \cite{movieclips}, which contains $\sim$30K videos from $\sim$3K movies. Each video is typically one to three minutes long. The benchmark contains nine tasks covering a wide range of long-form video understanding tasks. These $9$ tasks fall into three main categories: (i) \textbf{content understanding}, which consists of (\textit{`relationship', `speaking style', `scene/place'}) prediction, (ii) \textbf{metadata prediction}, which includes (\textit{`director', `genre', `writer', and `movie release year'}) classification, and (iii)  \textbf{user engagement}, which requires predicting (\textit{`YouTube like ratio', and `YouTube popularity'}). 

The content understanding and the metadata prediction tasks are evaluated using the standard top-1 accuracy metric, whereas the user engagement prediction tasks are evaluated using mean-squared error (MSE). Following~\cite{wu2021towards}, we use standard splits and train our model using video clips of 60 seconds.

We compare our proposed ViS4mer model with the previous methods validated on this benchmark. In particular, we use the same baselines as in~\cite{wu2021towards}. Additionally, we implement our own Long Sequence Transformer (\textbf{LST}) baseline, which follows exactly the same design as our ViS4mer model except for the S4 layers, which are replaced with standard self-attention layers.

We present our results in Table~\ref{tab:main_result} where we show that ViS4mer achieves state-of-the-art performance in most tasks. Specifically, ViS4mer outperforms both long-range video models (VideoBERT, and Object Transformer) in the content understanding and metadata prediction tasks by a significant margin and achieves comparable performance in the user engagement tasks. Furthermore, we also demonstrate that ViS4mer outperforms our Long Sequence Transformer baseline, which suggests the superiority of S4 layers over the standard self-attention layers for these tasks.

\begin{table}[t]
    \centering
    \caption{\small Comparison to prior works on the LVU dataset. Compared to previous long-range video models (VideoBERT, and Object Transformer), ViS4mer achieves significantly better accuracy in most tasks. Furthermore, ViS4mer also outperforms our implemented Long Sequence Transformer baseline, which uses the same design as our model, except for the S4 layers, which are replaced with the self-attention layers.}
    \label{tab:main_result}
    
    \resizebox{\textwidth}{!}{%
    \begin{tabular}{@{\extracolsep{2pt}} ll|lllllllll @{}}
    
      & Sequence &\multicolumn{3}{c}{Content ($\uparrow$)} & \multicolumn{4}{c}{Metadata ($\uparrow$)} & \multicolumn{2}{c}{User ($\downarrow$)}  \\
      
     \cline{3-5} \cline{6-9} \cline{10-11}
     
      & Model & Relation & Speak & Scene & Director & Genre & Writer & Year & Like & Views\\
      \hline
      
      SlowFast+NL\cite{feichtenhofer2019slowfast, wang2018non} & non-local & 52.40 & 35.80 & 54.70 & 44.90 & 53.00 & 36.30 & \textbf{52.50} &0.38 & 3.77 \\
      
      VideoBERT~\cite{sun2019videobert} & self-attention & 52.80 & 37.90 & 54.90 & 47.30 & 51.90 & 38.50 & 36.10 & 0.32 & 4.46 \\
      
      Obj. Transformer\cite{wu2021towards} & self-attention & 53.10 & 39.40 & 56.90 & 51.20 & 54.60 & 34.50 & 39.10 & \textbf{0.23} & \textbf{3.55}\\
     
      \hline
      
      Long Seq. Transformer & self-attention & 52.38&	37.31&	62.79&	56.07&	52.70&	42.26&	39.16&	0.31&	3.83 \\
      
      \textbf{ViS4mer} & state-space & \textbf{57.14} & \textbf{40.79} & \textbf{67.44} & \textbf{62.61} & \textbf{54.71} & \textbf{48.8} & \underline{44.75} & \underline{0.26} & \underline{3.63} \\
      \hline
    \end{tabular}}
    
\end{table}

\subsection{Ablation Studies}

\textbf{Detailed Comparison with Self-attention.} Since most previous methods predominantly use self-attention for long sequence video modeling, we compare our state-space (i.e., S4) based design with the equivalent self-attention-based approaches. For these comparisons, we use the same Long Sequence Transformer baseline (described in the previous subsection), which replaces all state-space layers of the ViS4mer model with the self-attention layers. 

Specifically, we compare the performance of ViS4mer and Long Sequence Transformer by varying the number of input tokens for both models. We use video clips of 60 seconds and a frame per second rate of 1. We vary the number of input tokens to $60, 1500, 2940$, and $11760$ by applying spatial max-pooling with a varying stride individually on each frame before feeding the tokens into the model. Note that the $60$ token baseline corresponds to a model that operates only on the frame-level CLS tokens. 

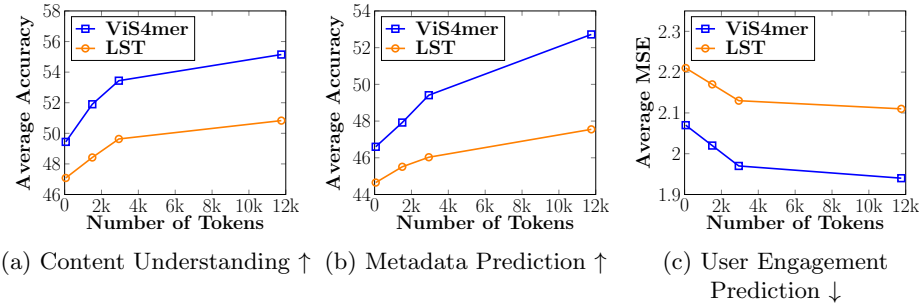
\begin{figure}[t]
    \centering
    \subfloat[\centering Content Understanding $\uparrow$]{{
    \resizebox{0.32\textwidth}{!}{%
    \begin{tikzpicture}[font={\LARGE}]
        \begin{axis}[
            xlabel={\textbf{Number of Tokens}},
            ylabel={\textbf{Average Accuracy}},
            xmin=0, xmax=12,
            xtick={0, 2, 4, 6, 8, 10, 12},
            xticklabels={0, 2k,4k, 6k, 8k, 10k, 12k},
            ymin=46, ymax=58,
            legend cell align={left},
            legend pos=north west
        ]
        
        \addplot[line width=.5mm, color=blue, mark=square, mark size=3pt]
            coordinates {
            (.06, 49.44)(1.5, 51.9)(2.94, 53.44)(11.76, 55.15)};\label{state-space}
        
        \addplot[line width=.5mm, color=orange,mark=o, mark size=3pt]
            coordinates {
            (.06, 47.08)(1.5, 48.42)(2.94, 49.63)(11.76, 50.83)};\label{self-attention}
            
       % \legend{\textbf{ state-space}, \textbf{ self-attention}}
       \legend{\textbf{ ViS4mer}, \textbf{ LST}}
        \end{axis}
        \end{tikzpicture}
        }
    }}%
    \subfloat[\centering Metadata Prediction $\uparrow$]{{
    \resizebox{0.32\textwidth}{!}{%
     \begin{tikzpicture}[font={\LARGE}]
        \begin{axis}[
            xlabel={\textbf{Number of Tokens}},
            ylabel={\textbf{Average Accuracy}},
            xmin=0, xmax=12,
            xtick={0, 2, 4, 6, 8, 10, 12},
            xticklabels={0, 2k,4k, 6k, 8k, 10k, 12k},
            ymin=44, ymax=54,
            legend cell align={left},
            legend pos=north west
        ]
        
        \addplot[ line width=.5mm, color=blue, mark=square, mark size=3pt]
            coordinates {
            (.06, 46.60)(1.5, 47.92)(2.94, 49.41)(11.76, 52.72)};\label{state-space}
        
        \addplot[line width=.5mm, color=orange,mark=o, mark size=3pt]
            coordinates {
            (.06, 44.65)(1.5, 45.51)(2.94, 46.03)(11.76, 47.55)};\label{self-attention}
            
        %\legend{\textbf{ state-space}, \textbf{ self-attention}}
        \legend{\textbf{ ViS4mer}, \textbf{ LST}}
        \end{axis}
        \end{tikzpicture}
        }
    }}%
    \subfloat[\centering User Engagement Prediction $\downarrow$]{{
    \resizebox{0.32\textwidth}{!}{%
     \begin{tikzpicture}[font={\LARGE}]
        \begin{axis}[
            xlabel={\textbf{Number of Tokens}},
            ylabel={\textbf{Average MSE}},
            xmin=0, xmax=12,
            xtick={0, 2, 4, 6, 8, 10, 12},
            xticklabels={0, 2k,4k, 6k, 8k, 10k, 12k},
            ymin=1.90, ymax=2.35,
            legend cell align={left},
            legend pos=north west
        ]
        
        \addplot[ line width=.5mm, color=blue, mark=square, mark size=3pt]
            coordinates {
            (.06, 2.07)(1.5, 2.02)(2.94, 1.97)(11.76, 1.94)};\label{state-space}
        
        \addplot[line width=.5mm, color=orange,mark=o, mark size=3pt]
            coordinates {
            (.06, 2.21)(1.5, 2.17)(2.94, 2.13)(11.76, 2.11)};\label{self-attention}
            
        %\legend{\textbf{ state-space}, \textbf{ self-attention}}
        \legend{\textbf{ ViS4mer}, \textbf{ LST}}
        \end{axis}
        \end{tikzpicture}
        }
    }}%
    \caption{\small We compare the performance of our ViS4mer and Long Sequence Transformer (LST) as a function of the number of input tokens on \textbf{(a)} the content understanding (using top-1 acc.), \textbf{(b)} the metadata prediction (using top-1 acc.), and \textbf{(c)} the user engagement prediction (using MSE) tasks. ViS4mer performs better for all number of tokens in all tasks.}%
    \label{fig:self-attention}%
\end{figure}

We present our results for this study in Figure~\ref{fig:self-attention}, where we plot (a) the average accuracy of three content understanding tasks, (b) the average accuracy of four metadata prediction tasks, and (c) the average MSE of two user engagement prediction tasks as a function of the number of input tokens. Based on these results, we observe that increasing the number of input tokens increases the performance of both ViS4mer and Long Sequence Transformer. We also note that ViS4mer achieves better performance than LST in all cases, which suggests that the state-space layers are superior to self-attention layers in this setting. Furthermore, our results indicate that the performance gap between the two methods increases as we increase the number of tokens. This observation suggests that the proposed ViS4mer architecture is more effective at incorporating information from very long video sequences. 

\begin{table}[t]
    \centering
    \caption{\small The GPU memory requirements (in GB) and the training speed (sample/seconds) of our state-space-based ViS4mer and the self-attention-based Long Sequence Transformer (LST). We compare both of these approaches while varying the number of input tokens. As we increase the number of tokens, the GPU memory and computation requirement of self-attention grows more rapidly for the LST baseline than for ViS4mer. Overall, ViS4mer requires $8\times$ less GPU  memory and is $2.63\times$ times faster than the self-attention baseline while operating on very long video sequences (i.e., $11,760$ spatiotemporal tokens).}
    \label{tab:training cost self-attention}
    \begin{tabular}{@{\extracolsep{2pt}}x{1.8cm}|x{2cm}x{2cm}|x{2cm}x{2cm}@{}}
        \# of Tokens & \multicolumn{2}{c}{Samples/s ($\uparrow$)} & \multicolumn{2}{c}{GPU Memory (GB)($\downarrow$)} \\
        \cline{1-5} %\cline{4-5}
        %\# of Tokens & state-space & self-attention & state-space & self-attention \\
         & ViS4mer & LST & ViS4mer & LST \\
        %\hline
        60 & \textbf{12.46} & 8.85 & \textbf{2.23} & 2.45\\
        %\hline
        1,500 & \textbf{8.27} & 6.31 & \textbf{3.61} & 3.99\\
        %\hline
        2,940 & \textbf{6.25} & 4.47 & \textbf{3.67} & 5.43\\
        %\hline
        11,760 & \textbf{4.95} & 1.88 & \textbf{5.15} & 41.38\\
        \hline
        
    \end{tabular}
\end{table}

\textbf{Computational Cost Analysis.} Additionally, in Table~\ref{tab:training cost self-attention}, we investigate the GPU memory requirements (in GB) and the training speed (sample/second) of ViS4mer and Long Sequence Transformer while varying the number of input tokens in the same way as was done in our previous analysis. These results indicate that in addition to being more accurate, ViS4mer is also significantly more memory-efficient and faster in all settings. It is worth mentioning, that when increasing the number of tokens to a large number ($11,760$) the memory requirement of the self-attention-based model grows very rapidly, requiring $41.38$GB of GPU memory. In contrast, ViS4mer is much more memory-efficient, requiring only $5.15$GB of GPU memory. Moreover, based on these results, we observe that ViS4mer is $2.63\times$ faster compared to Long Sequence Transformer when operating on sequences consisting of $11,760$ spatiotemporal tokens. 

\textbf{Comparison with Other Efficient Attention Schemes.} We also compare our method with other efficient self-attention schemes (e.g., Performer\cite{choromanski2020rethinking} and Orthoformer\cite{patrick2021keeping}) that do not require quadratic complexity with respect to the input length. We construct such models by replacing the state-space layers of the ViS4mer with the corresponding efficient self-attention layers and keeping all other settings the same as for our model. Table~\ref{tab:efficient attention} shows the results of these comparisons. We can observe that ViS4mer achieves the best performance in most LVU benchmark tasks while requiring similar memory and computation cost as other efficient attention schemes (e.g., Performer and Orthoformer).

\begin{table}[t]
    \centering
    \caption{\small Comparison with other efficient attention schemes. ViS4mer outperforms Performer\cite{choromanski2020rethinking} and Orthoformer\cite{patrick2021keeping} on the LVU benchmark while requiring similar memory and computation cost.}
    \label{tab:efficient attention}
    
    \resizebox{\textwidth}{!}{%
    \begin{tabular}{@{\extracolsep{1pt}} l|lllllllll|ll @{}}
    
     &\multicolumn{3}{c}{Content ($\uparrow$)} & \multicolumn{4}{c}{Metadata ($\uparrow$)} & \multicolumn{2}{c}{User ($\downarrow$)} && \\
      
    \cline{2-4} \cline{5-8} \cline{9-10}
     
      &Relation & Speak & Scene & Director & Genre & Writer & Year & Like & Views&Sam./s ($\uparrow$)&Mem ($\downarrow$)\\
      \hline
      
      Self-attention & 52.38&	37.31&	62.79&	56.07&	52.70&	42.26&	39.16&	0.31&	3.83 & 1.88 & 41.38\\
      
      Performer &50.00&	38.80&	60.46&	58.87&	49.45&	48.21&	41.25&	0.31&	3.93 & 4.67 & 5.93\\
      
      Orthoformer &50.00&	39.30&	66.27&	55.14&	\textbf{55.79}&	47.02&	43.35&	0.29&	3.86 &4.85& 5.56\\
      
      State-space & \bf 57.14 & \textbf{40.79} & \textbf{67.44}& \textbf{62.61} & 54.71 & \textbf{48.8} & \textbf{44.75} & \textbf{0.26} & \textbf{3.63} & \textbf{4.95} & \textbf{5.15}\\
      \hline
    \end{tabular}
    }
    
\end{table}

\begin{table}
    \centering
    \caption{\small Ablation on the ViS4mer architecture design. We observe that both (i) multi-scale feature learning (enabled by pooling) and (ii) progressive channel dimension reduction are critical for the best performance on the LVU benchmark. ViS4mer achieves substantially better performance compared to the vanilla S4 model while being $1.41\times$ faster and requiring $2.2\times$ less GPU memory.}
    \label{tab:ms4 architecture}
    
    \begin{tabular}{@{\extracolsep{2pt}}c c| lllll @{}}
    
    Pooling & Scaling & Content($\uparrow$) & Metadata($\uparrow$) & User($\downarrow$) & Samples/s($\uparrow$) & Memory(GB)($\downarrow$)\\
    \hline
    \ding{55} & \ding{55} & 49.53 & 49.26 & 2.30& 2.25 & 7.27 \\
    \ding{51} & \ding{55} & 48.96& 49.77 & 2.10 & 3.98 & 5.96\\
    \ding{55} & \ding{51} & 52.25 & 48.79 & 2.09 & 4.12 & 5.95\\
    \ding{51} & \ding{51} & \textbf{55.12}& \textbf{52.72}& \textbf{1.94} & \textbf{4.95}& \textbf{5.15}\\
    \hline
    \end{tabular}
    
\end{table}

\textbf{ViS4mer Architecture Analysis.} In Table~\ref{tab:ms4 architecture}, we analyze the significance of (i) multi-scale feature learning, which is enabled by the pooling layers, and (ii) the channel dimensionality reduction, which improves efficiency and reduces overfitting. These results indicate that both of these architecture design choices contribute not only to better performance on the LVU benchmark but also to the higher efficiency of ViS4mer. Specifically, compared to the vanilla S4 model, our final ViS4mer achieves $5.5\%$, and $3.5\%$ better performance on the content understanding and metadata prediction tasks respectively, and $0.36$ lower MSE on the user engagement tasks. Furthermore, ViS4mer has $2.2\times$ faster run-time and $1.41\times$ smaller GPU memory usage than the vanilla S4 model.

\textbf{Short-range Encoder Ablation.} To validate the robustness of our model, we also conduct experiments with different short-range encoder models. Specifically, we experiment with four popular short-range models, which includes both CNN and Transformer-based encoders: (i) SlowFast~\cite{feichtenhofer2019slowfast}, which was used by the previous Object Transformer method~\cite{wu2021towards}, (ii) ConvNeXt~\cite{liu2022convnet}, (iii) Swin Transformer~\cite{liu2021swin}, and (iv) ViT~\cite{dosovitskiy2020image}. We report our results of this analysis in Table~\ref{tab:backbone}. Based on these results, we first note that ViS4mer outperforms Object Transformer in $6$ tasks while using SlowFast~\cite{feichtenhofer2019slowfast}, and $9$ tasks while using ViT~\cite{dosovitskiy2020image}. Furthermore, we observe that using ViT as our short-range encoder leads to the best performance in $6$ out of $9$ LVU benchmark tasks.

\begin{table}[t]
    \centering
    \caption{\small Short-range encoder ablation. ViS4mer outperforms Object Transformer in 6 out of the 9 tasks while using the same short-range model (SlowFast~\cite{feichtenhofer2019slowfast}). Moreover, we observe that using ViT as our short-range encoder produced the best results in $6$ out of $9$ LVU benchmark tasks.}
    \label{tab:backbone}
    
    \resizebox{\textwidth}{!}{%
    \begin{tabular}{@{\extracolsep{2pt}} lllllllllll @{}}
    
      &&\multicolumn{3}{c}{Content ($\uparrow$)} & \multicolumn{4}{c}{Metadata ($\uparrow$)} & \multicolumn{2}{c}{User ($\downarrow$)}  \\
      
     \cline{3-5} \cline{6-9} \cline{10-11}
     
      Model& Encoder & Relation & Speak & Scene & Director & Genre & Writer & Year & Like & Views\\
      \hline
      
      \multirow{2}{*}{Obj. Trans.\cite{wu2021towards}}&
      SlowFast~\cite{feichtenhofer2019slowfast} & 53.10 & 39.40 & 56.90 & 51.20 & 54.60 & 34.50 & 39.10 & \textbf{0.23} & \textbf{3.55}\\&
      
      ViT~\cite{dosovitskiy2020image} & 54.76 & 33.17 & 52.94 & 47.66 & 52.74 & 36.30 & 37.76 & 0.30 & 3.68\\
      
      \hline
      
      \multirow{4}{*}{ViS4mer}&
      SlowFast~\cite{feichtenhofer2019slowfast} &\textbf{59.52}&	40.29&	60.46&	53.27&	52.74&	42.85&	39.86& 0.27& 3.70\\&
      
      ConvNeXt~\cite{liu2022convnet} &\textbf{59.52}&	38.30&	62.79&	57.00&	54.40&	45.83&	42.65& 0.30&	3.74\\&
      
      Swin~\cite{liu2021swin} & 54.76&	37.31&	61.62&	56.07&	49.45&	47.61&	39.86& 0.31&	3.56\\&
      
      ViT\cite{dosovitskiy2020image} & 57.14 & \textbf{40.79} & \textbf{67.44}& \textbf{62.61} & \textbf{54.71} & \textbf{48.8} & \textbf{44.75} & 0.26 & 3.63\\
      \hline
    \end{tabular}
    }
    
\end{table}

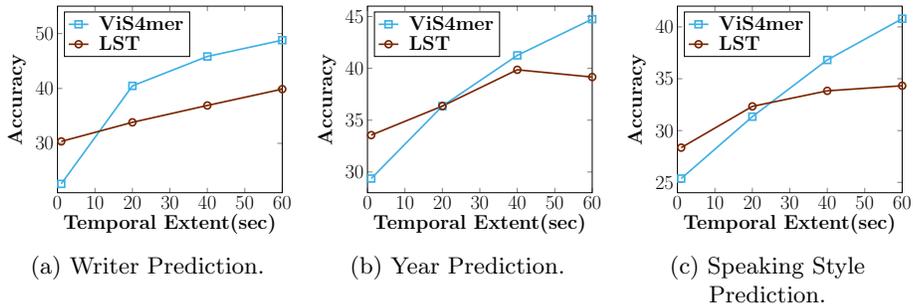
\begin{figure}[t]
    \centering
    \subfloat[\centering Writer Prediction.]{{
    \resizebox{0.32\textwidth}{!}{%
    \begin{tikzpicture}[font={\LARGE}]
        \begin{axis}[
            xlabel={\textbf{Temporal Extent(sec)}},
            ylabel={\textbf{Accuracy}},
            xmin=0, xmax=60,
            ymin=21, ymax=55,
            legend cell align={left},
            legend pos=north west
        ]
        \addplot[line width=.5mm, color=CornflowerBlue, mark=square, mark size=3pt]
            coordinates {
            (1, 22.61)(20, 40.47)(40, 45.83)(60, 48.8)};\label{state-space}
        
        \addplot[line width=.5mm, color=Brown,mark=o, mark size=3pt]
            coordinates {
            (1, 30.35)(20, 33.83)(40, 36.9)(60, 39.88)};\label{self-attention}
            
        %\legend{\textbf{ state-space}, \textbf{ self-attention}}
        \legend{\textbf{ ViS4mer}, \textbf{ LST}}
        \end{axis}
        \end{tikzpicture}
        }
    }}%
    \subfloat[\centering Year Prediction.]{{
    \resizebox{0.32\textwidth}{!}{%
    \begin{tikzpicture}[font={\LARGE}]
        \begin{axis}[
            xlabel={\textbf{Temporal Extent(sec)}},
            ylabel={\textbf{Accuracy}},
            xmin=0, xmax=60,
            ymin=28, ymax=46,
            legend cell align={left},
            legend pos=north west
        ]
        \addplot[line width=.5mm, color=CornflowerBlue, mark=square, mark size=3pt]
            coordinates {
            (1, 29.37)(20, 36.36)(40, 41.25)(60, 44.75)};\label{state-space}
        
        \addplot[line width=.5mm, color=Brown,mark=o, mark size=3pt]
            coordinates {
            (1, 33.56)(20, 36.36)(40, 39.86)(60, 39.16)};\label{self-attention}
            
        %\legend{\textbf{ state-space}, \textbf{ self-attention}}
        \legend{\textbf{ ViS4mer}, \textbf{ LST}}
        \end{axis}
        \end{tikzpicture}
        }
    }}%
    \subfloat[\centering Speaking Style Prediction.]{{
    \resizebox{0.32\textwidth}{!}{%
    \begin{tikzpicture}[font={\LARGE}]
        \begin{axis}[
            xlabel={\textbf{Temporal Extent(sec)}},
            ylabel={\textbf{Accuracy}},
            xmin=0, xmax=60,
            ymin=24, ymax=42,
            legend cell align={left},
            legend pos=north west
        ]
        \addplot[line width=.5mm, color=CornflowerBlue, mark=square, mark size=3pt]
            coordinates {
            (1, 25.37)(20, 31.34)(40, 36.81)(60, 40.79)};\label{state-space}
        
        \addplot[line width=.5mm, color=Brown, mark=o, mark size=3pt]
            coordinates {
            (1, 28.35)(20, 32.33)(40, 33.83)(60, 34.32)};\label{self-attention}
            
        %\legend{\textbf{ state-space}, \textbf{ self-attention}}
        \legend{\textbf{ ViS4mer}, \textbf{ LST}}
        \end{axis}
        \end{tikzpicture}
        }
    }}%
    \caption{\small Performance on the \textbf{(a)} Writer Prediction, \textbf{(b)} Year Prediction, and \textbf{(c)} Speaking Style Prediction tasks as a function of the input video duration. Based on these results, we observe that the Long Sequence Transformer (LST) performs better on very short clips. However, ViS4mer excels on much longer clips indicating its effectiveness at modeling long video sequences.}
    
    \label{fig:temporal}
\end{figure}

\textbf{Temporal Extent Ablation.} Additionally, we compare the long-range temporal reasoning abilities of our state-space-based ViS4mer and an equivalent self-attention-based Long Sequence Transformer. In particular, we train both of these models using video inputs spanning $1, 20, 40$, and $60$ seconds. In Figure~\ref{fig:temporal}, we illustrate our results on three LVU tasks (\textit{i.e.}, Writer prediction, Year prediction, and Speaking style prediction). Our results suggest that while the self-attention-based approach performs better when applied on clips that span short temporal extent (\textit{i.e.}, 1s in duration), the state-space model achieves much better performance when video inputs span long temporal extents (i.e., $40s$ and more). These results suggest that compared to self-attention, the state-space layers enable more effective long-range temporal reasoning. %over the video clips that span long temporal extents.

\begin{table}[t]
    \centering
    \caption{\small Evaluation on two long-range procedural activity classification datasets, \textit{i.e.}, COIN~\cite{tang2019coin, tang2020comprehensive}, and  Breakfast~\cite{kuehne2014language}. ViS4mer achieves comparable performance as the best performing, Distant Supervision~\cite{lin2022learning} framework, while using significantly less pretraining data. This indicates ViS4mer's ability to generalize to other domains.}
    \label{tab:other datasets}
    \begin{subtable}{1\linewidth}
      \centering
        \caption{Long-range procedural activity classification on the Breakfast~\cite{kuehne2014language} dataset.}
        \begin{tabular}{@{\extracolsep{2pt}} lccc@{}}
        Model & Pretraining Dataset & Pretraining Samples & Accuracy($\uparrow$) \\
        \hline
        VideoGraph~\cite{hussein2019videograph} & Kinetics-400 & 306K & 69.50 \\
        Timeception~\cite{hussein2019timeception} & Kinetics-400 & 306K & 71.30 \\
        GHRM~\cite{zhou2021graph} & Kinetics-400 & 306K & 75.50 \\
        Distant Supervision~\cite{lin2022learning} & HowTo100M & \bf 136M & \textbf{89.90} \\
        \textbf{ViS4mer} & Kinetics-600 & 495K & \underline{88.17} \\
        \hline
        \end{tabular}
    \end{subtable}%
    \hspace{0.5cm}
    \begin{subtable}{1\linewidth}
      \centering
        \caption{Long-range procedural activity classification on the COIN~\cite{tang2019coin} dataset.}
        \begin{tabular}{@{\extracolsep{2pt}}lccc@{}}
        Model & Pretraining Dataset & Pretraining Samples & Accuracy($\uparrow$) \\
        \hline
        TSN~\cite{tang2020comprehensive} & Kinetics-400 & 306K & 73.40 \\
        Distant Supervision~\cite{lin2022learning} & HowTo100M & \bf 136M & \textbf{90.00} \\
        \textbf{ViS4mer} & Kinetics-600 & 495K & \underline{88.41} \\
        \hline
        \end{tabular}
    \end{subtable} 
\end{table}

\subsection{Evaluation on Other Datasets}

Lastly, to evaluate ViS4mer's ability to generalize to other domains, we conduct experiments on two long-range procedural activity classification datasets:  Breakfast~\cite{kuehne2014language} and COIN~\cite{tang2019coin, tang2020comprehensive}. The Breakfast dataset contains $1,712$ videos of $10$ complex cooking activities. The average duration of the videos is $~2.32$ minutes. The COIN dataset consists of $11,827$ videos capturing $180$ diverse procedural tasks. The average length of a video is $~2.36$ minutes. 

Given the long duration of the videos in both datasets, we believe that these datasets are well suited to test our model's ability for long-range activity understanding. For all of our experiments, we use standard splits~\cite{hussein2019videograph, tang2020comprehensive} and measure performance in terms of activity classification accuracy. 

We report our results on these two datasets in Table~\ref{tab:other datasets}. Based on these results, we observe that ViS4mer outperforms previous approaches pretrained on Kinetics~\cite{kay2017kinetics, carreira2018short}. Moreover, ViS4mer achieves competitive performance as the state-of-the-art Distant Supervision method~\cite{lin2022learning}, which uses several orders of magnitude more pretraining data (i.e., HowTo100M~\cite{miech2019howto100m}). Thus, we believe that these results provide sufficient evidence of ViS4mer's generalization ability.

\subsection{Qualitative Results}

\begin{figure}
    \centering
    \includegraphics[width=\textwidth]{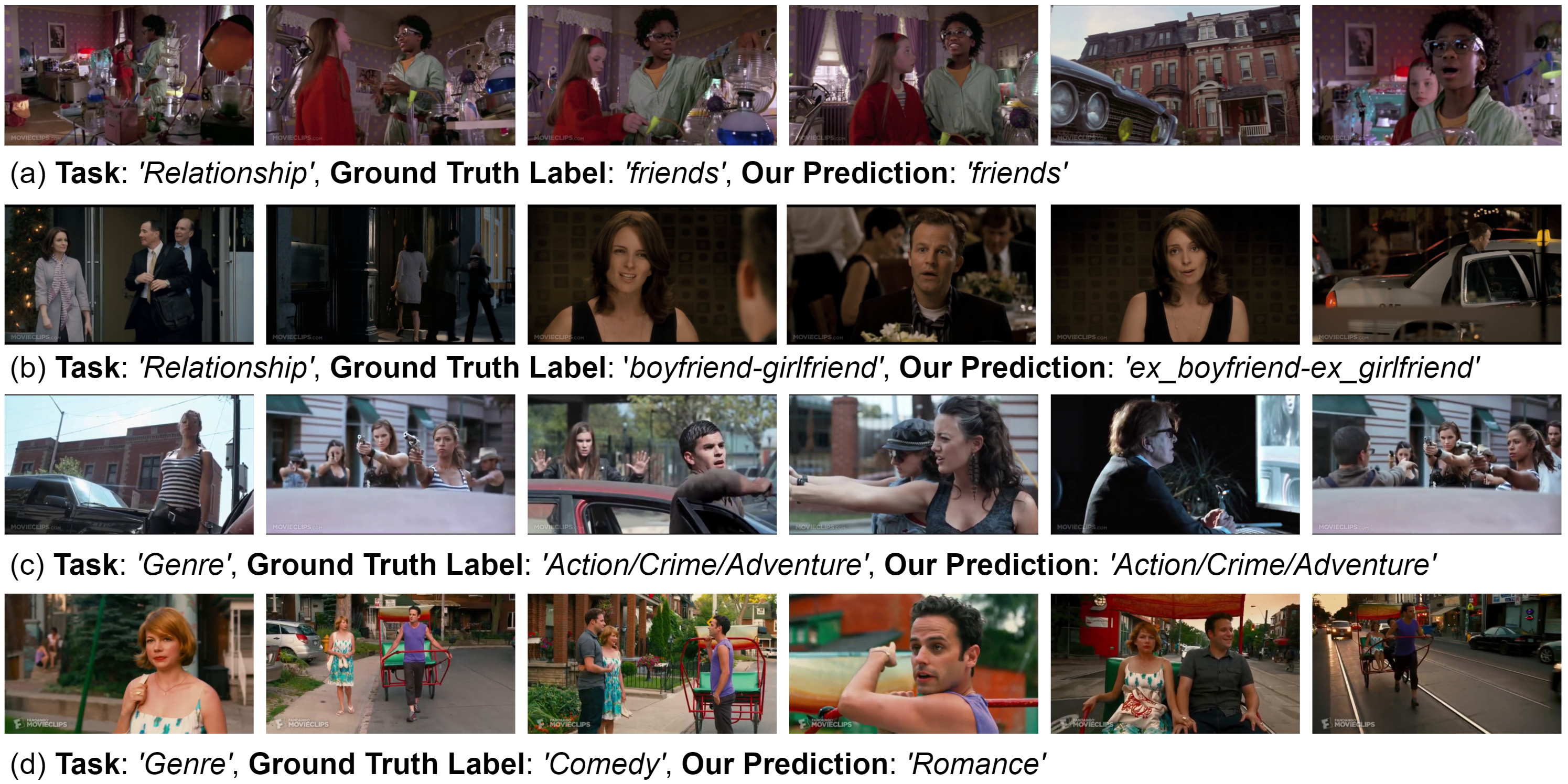}
    \caption{Our qualitative results on the LVU dataset. ViS4mer can effectively identify \textbf{(a)} the relationship among the characters and \textbf{(c)} the genre of the movie. Furthermore, the complex examples shown in the rows \textbf{(b)} \textbf{(d)} illustrate the difficulties of long-range movie understanding tasks.}
\label{fig:example}
\end{figure}

In Figure~\ref{fig:example}, we also illustrate some qualitative results on the LVU dataset. In particular, we demonstrate several instances of the correct and incorrect predictions of our ViS4mer model on the relationship and genre prediction tasks. These results indicate that our method can effectively identify the relationships among the characters (Figure~\ref{fig:example}(a)) and the genre of the movie (Figure~\ref{fig:example}(c)). Furthermore, these qualitative examples highlight some movie instances which are difficult to classify even for a human (Figures~\ref{fig:example}(b), (d)), thus, illustrating the challenging nature of long-range movie understanding tasks.

\section{Conclusion}

Combining the strength of self-attention and structured state-space sequence models, we introduce ViS4mer, an efficient framework for long-range movie video classification. Our method (i) is conceptually simple, (ii) achieves state-of-the-art results on several complex movie understanding tasks, (iii) has low memory requirement and computation cost, and (iv) successfully generalizes to other domains such as procedural activity classification. In the future, we plan to extend our work to other long-range video understanding tasks such as video summarization, question answering, and video grounding. 

\bibliographystyle{splncs04}
\bibliography{egbib}
\end{document}